# Image Denoising via the Adaptive Rank-Cluster Filter

Dmitry Pozdnyakov

*Oxagile LLC, Minsk, Belarus*
*E-mail: dr.dmitry.pozdnyakov@gmail.com; dmitry.pozdnyakov@oxagile.com*

**Abstract:** A spatial-local image-denoising filter is proposed, and its performance metrics are evaluated in comparison with baseline filtering algorithms, including the median, adaptive median, Gaussian, bilateral, Wiener, anisotropic diffusion, and non-local means. The developed filter is based on aligning the intensity value of the central pixel in a 3×3 window with the statistical majority intensity of one of the two clusters formed by optimal Otsu's partitioning of a pixel set sorted by intensity and trimmed to seven elements. This is followed by a fuzzy fusion of the calculated value with the median intensity of the pixels within the window. The proposed filter demonstrates the highest robustness to variations in image noise levels, particularly when processing mixed noise consisting of salt-and-pepper impulse noise and additive Gaussian noise in various proportions.

## Introduction

To date, deep learning methods have become the standard for developing noise filters for images and video frames [1–4], achieving the highest quality of image restoration. Nevertheless, interest in heuristic approaches for image and video denoising algorithms has declined but not disappeared [5–8], and there are highly compelling reasons for this. First, classical filtration methods possess fundamentally lower computational complexity, require no pre-training, and are entirely free from the "hallucination" effect (artifacts where non-existent details are generated) inherent to deep neural networks. This makes them indispensable for real-time systems and resource-constrained mobile platforms. Second, these methods demonstrate high stability and robustness when processing rare, atypical, or combined types of noise. While the performance of a neural network critically depends on the representativeness of the training dataset and drops sharply when encountering out-of-distribution noise, spatial-local filters, for instance, rely on the fundamental statistical properties of images. This allows them to effectively suppress even complex noise compositions without the risk of performance degradation. Third, unlike deep neural networks, which represent a "black box" with hard-to-predict stochastic behavior, classical filtration methods offer rigorous mathematical justification and deterministic properties. This guarantees complete predictability of their performance at object boundaries during sharp intensity transitions. In such regions, neural network models sometimes either blur textures or generate false contours, whereas spatial filters predictably control edge geometry. However, most existing modern spatial filters were originally designed to suppress a single, specific type of noise, demonstrating outstanding restoration quality for those particular cases. Yet, under mixed noise conditions—such as a combination of salt-and-pepper impulse noise and additive Gaussian noise—the denoising efficiency of these filters drops drastically.

Consequently, the purpose of this work is to develop and evaluate a spatial-local Adaptive Rank-Cluster filter (ARC-filter) intended for efficient mixed noise suppression in images. The filter is designed to preserve object contours with high precision while maintaining acceptable computational complexity, which is particularly critical for processing medical diagnostic images (such as X-ray, ultrasound, etc.). To achieve this, the following tasks are addressed: a mathematical framework for the filter algorithm is presented, and a comparative analysis of image restoration quality is performed, evaluating the proposed filter against several classical benchmarks using multiple key performance metrics on corrupted images.

## Theoretical background

We limit our consideration to low-intensity impulse noise of the salt-and-pepper type, when the fraction of corrupted pixels $\omega$ does not exceed 0.02. We also constrain the standard deviation of the additive Gaussian noise to $\eta \leq 0.2$ (for pixel intensities normalized to the range [0, 1]). The analysis of images heavily corrupted by salt-and-pepper noise [9] is primarily of methodological interest [5, 6, 9], whereas such high noise levels are rarely encountered in practical, real-world applications. This is due to the fact that modern technologies enable the manufacturing of high-quality digital electromagnetic radiation detectors with only isolated matrix structure defects (including dead pixels). Consequently, if a certain amount of impulse noise is present in the images, its origin is primarily linked to Gaussian noise caused by the "clipping" of the pixel intensity distribution tails at the boundaries of the detector's dynamic range under high signal amplification and/or low-light conditions.

Let us consider a sliding window with a minimal neighborhood around the analyzed central pixel $S_c$, specifically a 3×3 window. Given that $\omega \le 0.02$, it is highly improbable that the number of pixels corrupted by impulse noise within any such window will exceed two, even in the presence of Gaussian noise. Consequently, at each position of the 3×3 sliding window, by discarding the first and last pixels from the rank-ordered intensity set of nine pixels $S_1, \ldots, S_9$, it can be assumed with a very high probability that within this small local image area of the seven remaining pixels, either a single sharp edge exists—dividing the pixels into two intensity clusters characterized by a bimodal distribution—or the pixel intensities follow a unimodal distribution. If a unimodal distribution occurs, depending on its specific type, the optimal smoothing filter for the seven pixels could be the mean, the median (such as Rank-Ordered-Mean, Rank-Selection-Median, etc.) [10–12], or some intermediate estimator generally calculated as a weighted average. However, subsequent analysis of artifact generation during the processing of test images uncorrupted by additive Gaussian noise demonstrated that the median estimator of the local unimodal distribution ($S_5$) yields the minimum processing errors.

In the case of a bimodal distribution within the rank-ordered set of seven pixels, it is first necessary to partition them into two clusters in an optimal manner. This is straightforwardly achieved using the standard Otsu's method, which, in the scenario under consideration, yields the same result as $K$-means clustering for $K = 2$. To accomplish this, all six possible partitions—ranging from a "1 + 6" to a "6 + 1" element split between the clusters—are iteratively evaluated to find the specific partition that maximizes the between-class variance

$$\lambda_1 = \arg\max_{k=1..6} \left( \frac{1}{k(7-k)} \left[ 7\sum_{p=1}^{k} S_{p+1} - k\sum_{p=1}^{7} S_{p+1} \right]^2 \right), \quad (1)$$

where $\lambda_1$ is the number of elements in the cluster with lower intensity, and $\lambda_2 = 7 - \lambda_1$ is the number of elements in the cluster with higher intensity. Following this, the necessary statistical estimators are computed for the resulting clusters:

$$m_1 = \lambda_1^{-1} \sum_{p=1}^{\lambda_1} S_{p+1}, \quad (2)$$

$$m_2 = \lambda_2^{-1} \sum_{p=\lambda_1+1}^{7} S_{p+1}, \quad (3)$$

$$\eta_1 = \max\left( \varepsilon, \left[ \lambda_1^{-1} \sum_{p=1}^{\lambda_1} (S_{p+1} - m_1)^2 \right]^{1/2} \right), \quad (4)$$

$$\eta_2 = \max\left( \varepsilon, \left[ \lambda_2^{-1} \sum_{p=\lambda_1+1}^{7} (S_{p+1} - m_2)^2 \right]^{1/2} \right), \quad (5)$$

where $\varepsilon = 10^{-16}$ is a constant (machine epsilon) introduced to avoid singularities in subsequent computations. Additionally, the values of auxiliary parameters are calculated as follows:

$$\mu = \max\left(1, (m_2 - m_1)/(\eta_1 + \eta_2)\right), \quad (6)$$

$$S_{th} = (m_1\eta_2 + m_2\eta_1)/(\eta_1 + \eta_2), \quad (7)$$

where $S_{th}$ is the threshold level for cluster separation.

Since, in addition to impulse noise, additive Gaussian noise is explicitly considered, a transfer function in the form of two smoothly jointed error functions should be utilized to determine the optimally filtered value of $S_c$ in the two-cluster mode:

$$S_f = \begin{cases} S_{th} + \mu\eta_1 \operatorname{erf}\left(\dfrac{S_c - S_{th}}{\eta_1\sqrt{2\pi}}\right), & S_c < S_{th}; \\ S_{th} + \mu\eta_2 \operatorname{erf}\left(\dfrac{S_c - S_{th}}{\eta_2\sqrt{2\pi}}\right), & S_c \geq S_{th}. \end{cases} \quad (8)$$

For $\mu \gg 1$, this transfer function maps $S_c$ ($S_f \in (m_1, m_2)$) to either $m_1$ or $m_2$, depending on its membership in the first or second cluster, respectively. Conversely, as $\mu \to 1$, the value of $S_c$ is pulled toward $S_{th}$ for pixel intensities originating from the inner regions of the clusters, while its intensity values from the outer regions are bounded by $m_1$ and $m_2$ (representing a soft clipping of the outer tails of the pixel intensity distribution).

The final intensity transformation of the central pixel involves a fuzzy fusion of the filtration results obtained from the unimodal and bimodal modes, determined by the prominence of the two modes within the local pixel intensity distribution:

$$S_{res} = S_5 + (S_f - S_5)\operatorname{erf}\left(\frac{\mu - 1}{\sqrt{2}}\right). \quad (9)$$

## Results and discussion

Next, we evaluate and compare the noise suppression performance on four test images across several metrics using the following filters: a median filter (M) with a 3×3 window; an adaptive median filter (AM) with an initial 3×3 window expandable up to 9×9; a Gaussian filter (G) with a 5×5 window at $\sigma = 1$; a bilateral filter (B) with a 5×5 window at $\sigma = 1$ and a photometric (range) standard deviation of 1/3; a Wiener filter (W) with a 3×3 window; an anisotropic diffusion filter (AD) configured with 6 iterations, a gradient threshold of 0.1, and an exponential conduction function; a non-local means filter (NLM) utilizing a 3×3 similarity window within a 15×15 search window; and the proposed Adaptive Rank-Cluster filter (ARC).

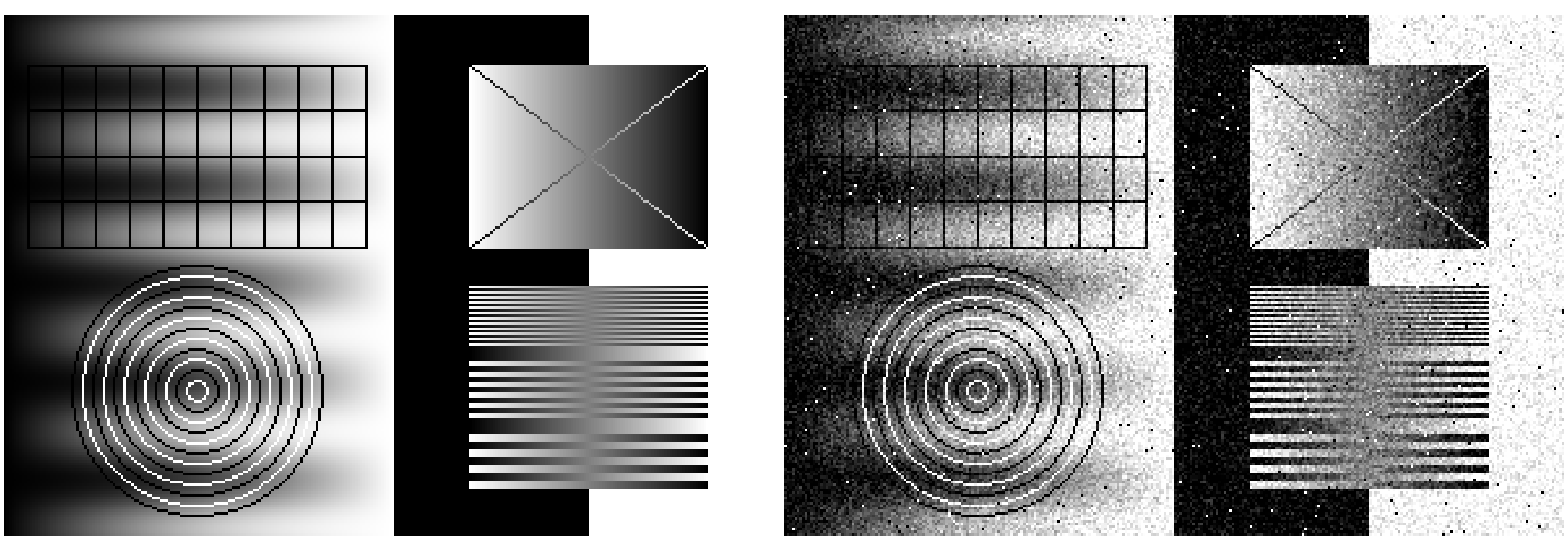

**Fig.1.** Synthetic 300×200 pixel image generated for evaluating the performance of noise filters: original image (left); example of a noisy image (N) at $\eta = 0.1$ and $\omega = 0.01$ (right)

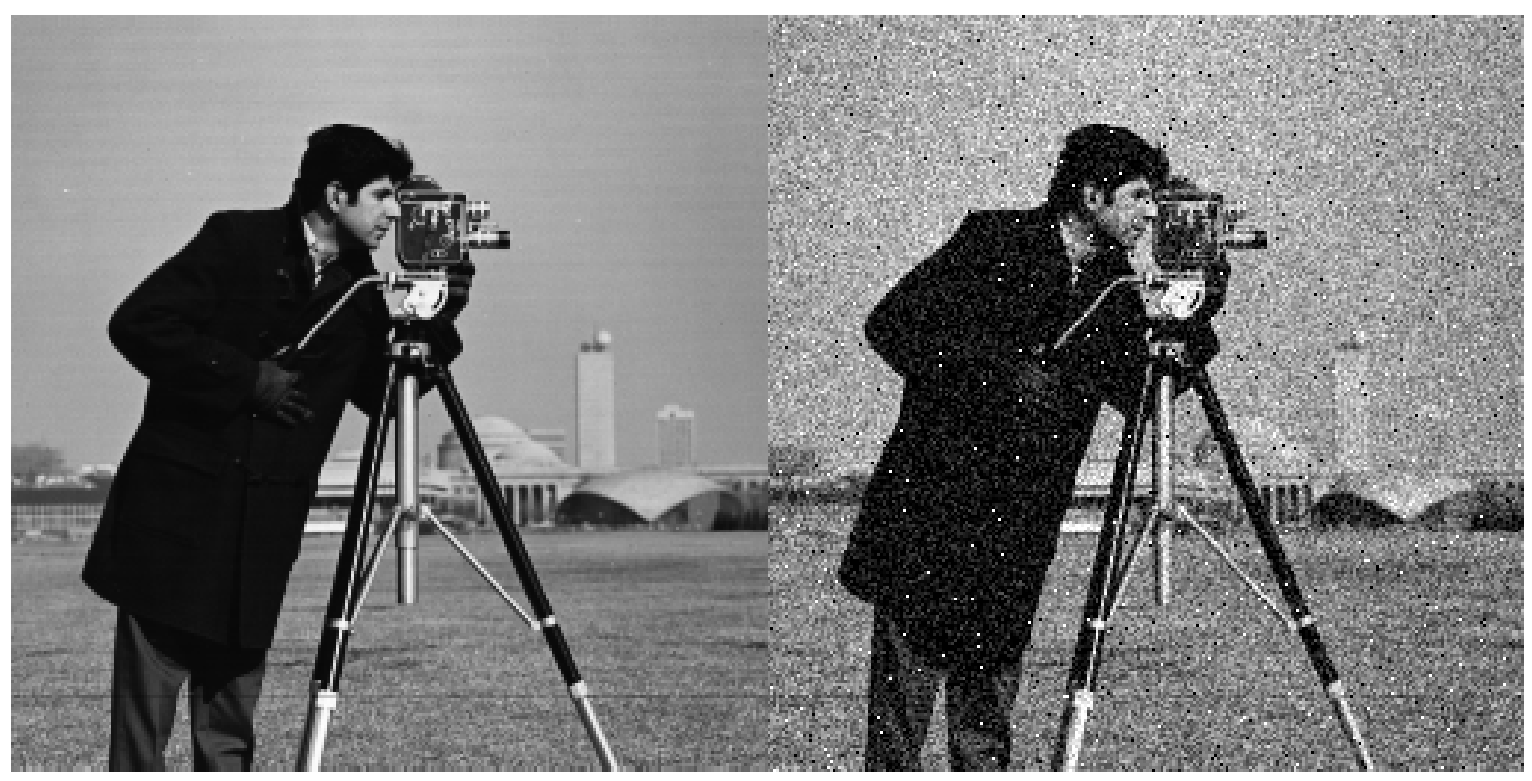

**Fig.2.** Standard "Cameraman" test image of 256×256 pixels: original image (left); example of a noisy image (N) at $\eta = 0.1$ and $\omega = 0.01$ (right)

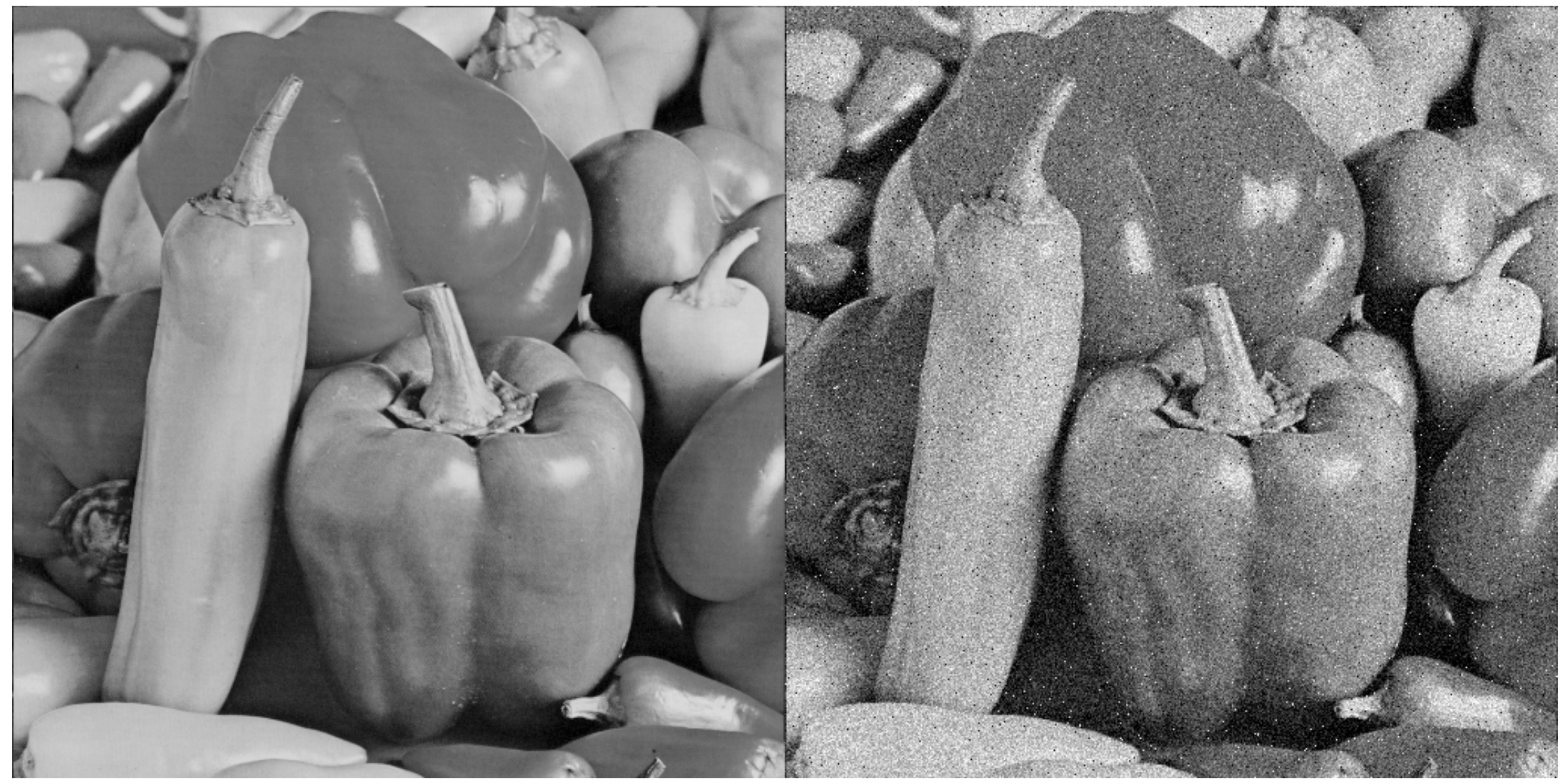

**Fig.3.** Standard "Peppers" test image of 512×512 pixels: original image (left); example of a noisy image (N) at $\eta = 0.1$ and $\omega = 0.01$ (right)

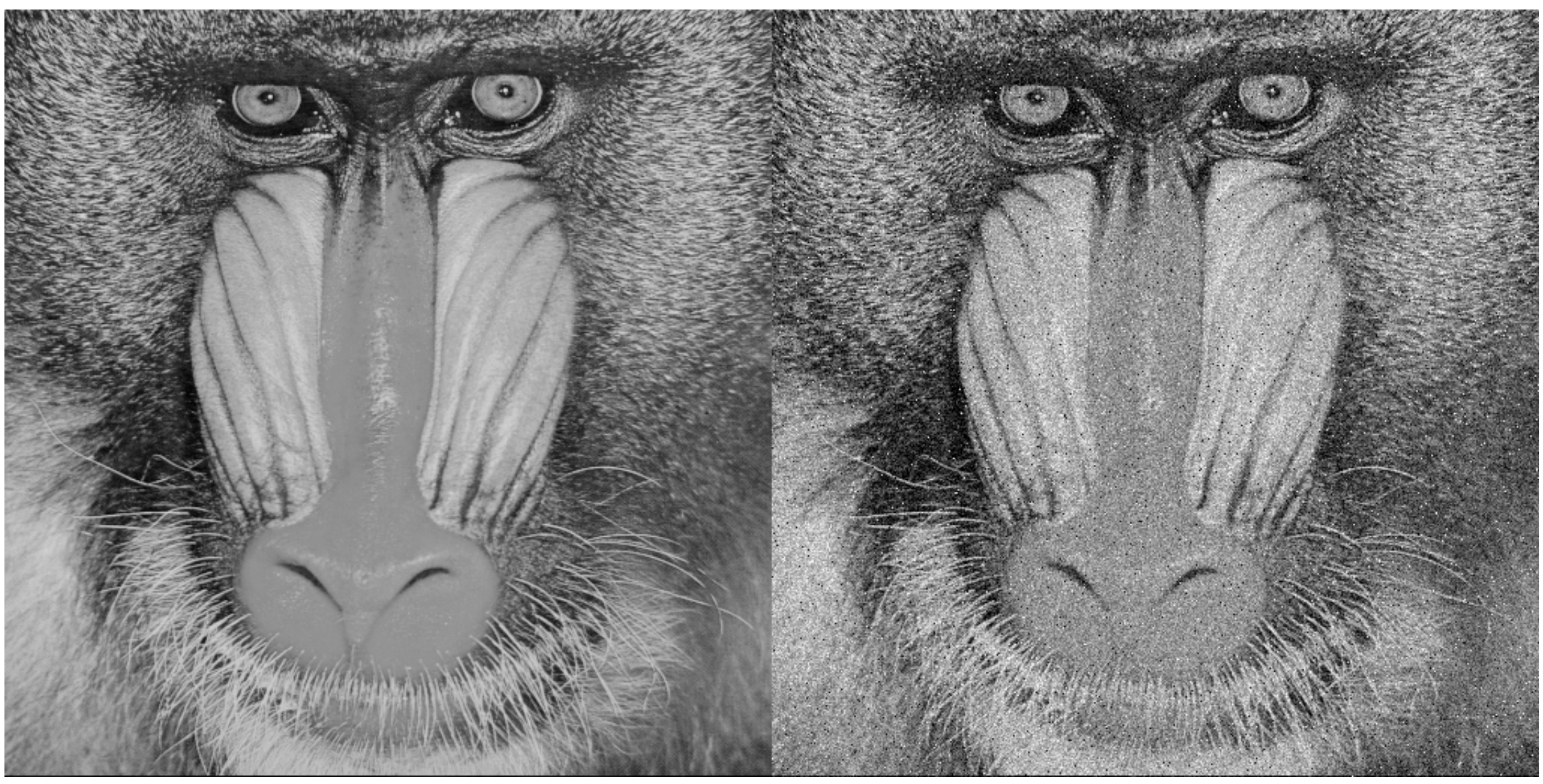

**Fig.4.** Standard "Baboon" test image of 512×512 pixels: original image (left); example of a noisy image (N) at $\eta = 0.1$ and $\omega = 0.01$ (right)

It should be noted that to ensure the consistency of metric evaluations, all computations were performed in the MATLAB environment using built-in functions and procedures. The algorithms for the AM- and ARC-filter were implemented in a manner analogous to built-in routines, featuring full vectorization and matrix calculation optimization. The metrics were obtained by averaging over 1000 distinct noise realizations on the test images.

The evaluation metrics included, first, the noise suppression factor $C_{NR}$ (expressed in dB), which is based on the mean squared error (MSE) and measures the variance deviation [9].

$$C_{NR} = 10 \cdot \lg\left( \sum_{ij} \left( S_{ij}^{(\mathrm{N})} - S_{ij}^{(0)} \right)^2 \Big/ \sum_{ij} \left( S_{ij}^{(\mathrm{F})} - S_{ij}^{(0)} \right)^2 \right), \quad (10)$$

where $i$ and $j$ represent the row and column indices of the image pixels $\mathbf{S} = \{S_{ij}\}$, and the subscripts “0”, “N”, and “F” denote the original, noisy, and filtered images, respectively. Second, the variation suppression factor

$$C_{VR} = 20 \cdot \lg\left( \sum_{ij} \left| S_{ij}^{(\mathrm{N})} - S_{ij}^{(0)} \right| \Big/ \sum_{ij} \left| S_{ij}^{(\mathrm{F})} - S_{ij}^{(0)} \right| \right) \quad (11)$$

(expressed in dB), calculated through the absolute deviation value, was evaluated based on the mean absolute error (MAE) [9]. Third, the structural similarity index $C_{SS}$ (SSIM [13]) was considered. Fourth, the relative computational efficiency with respect to the median filter (expressed in %) was evaluated:

$$C_{CE} = 100 \cdot T_M / T, \quad (12)$$

where $T_M$ is the processing time of the image by the median filter, and $T$ is the processing time by the filter under consideration.

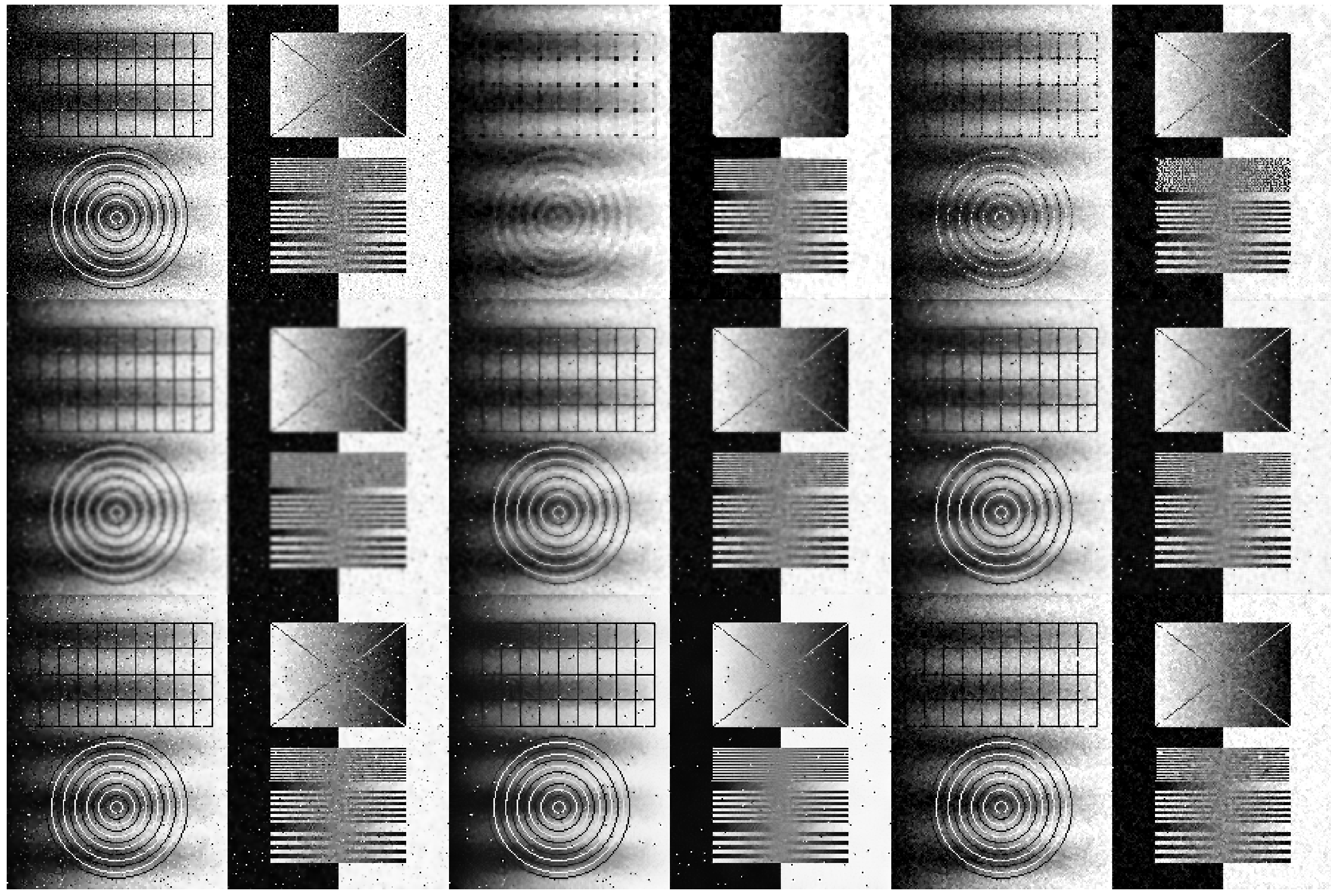

**Fig.5.** Processing results for the synthetic test image corrupted by mixed noise at $\eta = 0.1$ and $\omega = 0.01$: N, M, AM; G, B, W; AD, NLM, and ARC (ordered from left to right, top to bottom)

**Table 1a.** Average $C_{NR}$ (dB) for the noisy synthetic image processed by various filters

| $\eta$ | $\omega$ | M | AM | G | B | W | AD | NLM | ARC |
|---|---|---|---|---|---|---|---|---|---|
| $10^{-3}$ | 0.00 | – 46.538 | – 48.270 | – 44.024 | – 39.769 | – 35.854 | *– 22.446* | **– 14.461** | ***– 15.196*** |
| | 0.01 | – 8.6007 | – 10.331 | – 6.2042 | – 2.4224 | – 0.8471 | *– 0.0802* | ***– 0.0193*** | **7.0247** |
| | 0.02 | – 5.5959 | – 7.3194 | – 3.3145 | *0.0457* | ***0.5526*** | – 0.0200 | 0.0264 | **6.4301** |
| 0.05 | 0.00 | – 11.468 | – 7.7361 | – 9.2762 | – 5.4311 | – 2.2112 | **6.4579** | ***5.6980*** | *3.3365* |
| | 0.01 | – 6.6775 | – 2.9903 | – 4.6124 | – 1.1661 | 0.2055 | ***1.3218*** | *1.3113* | **5.4566** |
| | 0.02 | – 4.4660 | – 0.8244 | – 2.5248 | 0.5615 | ***1.0078*** | 0.7595 | *0.8904* | **5.6269** |
| 0.10 | 0.00 | – 5.5439 | – 3.8259 | – 3.7751 | – 0.5128 | 1.5034 | ***3.0004*** | **5.9451** | *2.9299* |
| | 0.01 | – 3.7191 | – 2.0248 | – 2.0776 | 0.8729 | *1.7544* | 1.7170 | ***3.3672*** | **3.9252** |
| | 0.02 | – 2.4504 | – 0.7772 | – 0.9290 | 1.7367 | *1.9657* | 1.2056 | ***2.8421*** | **4.3251** |
| 0.15 | 0.00 | – 2.2897 | – 2.4348 | – 0.8762 | 1.7556 | ***2.8002*** | 1.1807 | **5.8872** | *2.5646* |
| | 0.01 | – 1.3774 | – 1.5178 | – 0.0776 | 2.3209 | *2.7289* | 0.9201 | **5.2399** | ***3.0915*** |
| | 0.02 | – 0.6279 | – 0.7639 | 0.5589 | *2.7400* | 2.7366 | 0.7516 | **5.0131** | ***3.4043*** |
| 0.20 | 0.00 | – 0.2264 | – 1.8191 | 0.9186 | *2.9439* | ***3.3036*** | 0.5690 | **5.7090** | 2.4126 |
| | 0.01 | 0.3040 | – 1.2683 | 1.3482 | *3.2010* | ***3.2308*** | 0.4928 | **5.6419** | 2.7146 |
| | 0.02 | 0.7658 | – 0.7801 | 1.7149 | ***3.4046*** | *3.2059* | 0.4340 | **5.5926** | 2.9231 |

**Table 1b.** Average $C_{VR}$ (dB) for the noisy synthetic image processed by various filters

| $\eta$ | $\omega$ | M | AM | G | B | W | AD | NLM | ARC |
|---|---|---|---|---|---|---|---|---|---|
| $10^{-3}$ | 0.00 | – 49.210 | – 50.870 | – 51.479 | – 47.233 | – 43.626 | *– 28.854* | ***– 23.160*** | **– 20.513** |
| | 0.01 | – 19.199 | – 20.855 | – 21.926 | – 17.801 | – 15.443 | *– 5.2162* | ***– 4.9493*** | **5.1243** |
| | 0.02 | – 13.334 | – 14.981 | – 16.478 | –12.469 | – 11.023 | ***– 2.9731*** | *– 3.7591* | **7.4606** |
| 0.05 | 0.00 | – 5.3153 | – 4.3915 | – 6.7791 | – 3.4726 | – 1.1662 | **6.0809** | ***4.8587*** | *3.2099* |
| | 0.01 | – 4.1538 | – 3.2306 | – 5.9561 | – 2.6964 | – 1.1595 | **4.9822** | *3.8633* | ***3.9388*** |
| | 0.02 | – 3.1449 | – 2.2213 | – 5.2638 | – 2.0550 | – 1.1214 | ***4.2175*** | *3.1876* | **4.4500** |
| 0.10 | 0.00 | – 1.2305 | – 3.1182 | – 2.2533 | 0.2926 | 1.6070 | ***3.1385*** | **5.0788** | *2.8600* |
| | 0.01 | – 0.6727 | – 2.5361 | – 1.9533 | 0.5733 | 1.4238 | *2.8472* | **4.6804** | ***3.2040*** |
| | 0.02 | – 0.1614 | – 1.9932 | – 1.6825 | 0.8209 | 1.2924 | *2.5974* | **4.4146** | ***3.4730*** |
| 0.15 | 0.00 | 0.6739 | – 2.6928 | – 0.1403 | 1.8082 | *2.4748* | 1.2973 | **4.9643** | ***2.5103*** |
| | 0.01 | 1.0109 | – 2.3112 | 0.0012 | 1.9406 | *2.3233* | 1.2186 | **4.8876** | ***2.7167*** |
| | 0.02 | 1.3344 | – 1.9452 | 0.1343 | 2.0618 | *2.2030* | 1.1464 | **4.8419** | ***2.8846*** |
| 0.20 | 0.00 | 1.7773 | – 2.4941 | 1.0387 | *2.5073* | ***2.7173*** | 0.6519 | **4.7691** | 2.3156 |
| | 0.01 | 2.0063 | – 2.2164 | 1.1150 | *2.5785* | ***2.6234*** | 0.6218 | **4.7801** | 2.4493 |
| | 0.02 | 2.2216 | – 1.9459 | 1.1846 | ***2.6412*** | 2.5425 | 0.5935 | **4.7748** | *2.5607* |

**Table 1c.** Average $C_{SS}$ (SSIM) for the noisy synthetic image processed by various filters

| $\eta$ | $\omega$ | M | AM | G | B | W | AD | NLM | ARC |
|---|---|---|---|---|---|---|---|---|---|
| $10^{-3}$ | 0.00 | 0.6711 | 0.6269 | 0.7706 | 0.8897 | 0.9159 | *0.9838* | ***0.9961*** | **0.9992** |
| | 0.01 | 0.6711 | 0.6266 | 0.7106 | 0.8138 | 0.8074 | *0.8412* | ***0.8493*** | **0.9868** |
| | 0.02 | 0.6709 | 0.6262 | 0.6592 | ***0.7496*** | 0.7237 | 0.7328 | *0.7404* | **0.9672** |
| 0.05 | 0.00 | 0.5791 | 0.5788 | 0.6560 | 0.7668 | 0.7845 | **0.8821** | ***0.8749*** | *0.7873* |
| | 0.01 | 0.5775 | 0.5780 | 0.6156 | 0.7122 | 0.7043 | *0.7645* | ***0.7663*** | **0.7749** |
| | 0.02 | 0.5759 | 0.5771 | 0.5808 | 0.6656 | 0.6420 | *0.6748* | ***0.6853*** | **0.7588** |
| 0.10 | 0.00 | 0.4687 | 0.4215 | 0.5735 | *0.6752* | ***0.6786*** | 0.6575 | **0.7891** | 0.6090 |
| | 0.01 | 0.4664 | 0.4207 | 0.5449 | ***0.6340*** | *0.6192* | 0.5959 | **0.7123** | 0.5991 |
| | 0.02 | 0.4637 | 0.4198 | 0.5193 | ***0.5978*** | 0.5719 | 0.5458 | **0.6547** | *0.5874* |
| 0.15 | 0.00 | 0.3891 | 0.3351 | 0.5017 | ***0.5893*** | *0.5874* | 0.4659 | **0.7069** | 0.4941 |
| | 0.01 | 0.3863 | 0.3343 | 0.4810 | ***0.5585*** | *0.5440* | 0.4413 | **0.6610** | 0.4863 |
| | 0.02 | 0.3837 | 0.3335 | 0.4626 | ***0.5313*** | *0.5089* | 0.4196 | **0.6255** | 0.4775 |
| 0.20 | 0.00 | 0.3331 | 0.2750 | 0.4404 | ***0.5104*** | *0.5056* | 0.3756 | **0.6290** | 0.4135 |
| | 0.01 | 0.3305 | 0.2743 | 0.4259 | ***0.4883*** | *0.4759* | 0.3621 | **0.6024** | 0.4072 |
| | 0.02 | 0.3276 | 0.2736 | 0.4123 | ***0.4680*** | *0.4507* | 0.3497 | **0.5778** | 0.4003 |

**Table 1d.** Average $C_{CE}$ for the noisy synthetic image processed by various filters

| $\eta$ | $\omega$ | M | AM | G | B | W | AD | NLM | ARC |
|---|---|---|---|---|---|---|---|---|---|
| $10^{-3}$ | 0.00 | ***100.00*** | 4.3339 | **155.17** | 11.967 | *98.107* | 16.185 | 1.4003 | 6.0018 |
| | 0.01 | *100.00* | 4.2128 | **147.93** | 11.827 | ***104.08*** | 16.174 | 1.5479 | 5.9105 |
| | 0.02 | *100.00* | 4.2072 | **155.82** | 12.077 | ***106.44*** | 16.543 | 1.5552 | 6.1250 |
| 0.05 | 0.00 | *100.00* | 8.3183 | **280.76** | 18.525 | ***190.32*** | 27.881 | 2.6630 | 9.9400 |
| | 0.01 | *100.00* | 8.1932 | **267.73** | 18.643 | ***187.47*** | 27.927 | 2.6995 | 9.9240 |
| | 0.02 | *100.00* | 7.7440 | **270.26** | 18.888 | ***182.92*** | 27.821 | 2.7336 | 9.9784 |
| 0.10 | 0.00 | *100.00* | 8.3182 | **269.91** | 19.351 | ***185.42*** | 28.418 | 2.7797 | 9.7956 |
| | 0.01 | *100.00* | 8.1009 | **277.08** | 19.152 | ***193.86*** | 28.744 | 2.7842 | 10.044 |
| | 0.02 | *100.00* | 7.8527 | **271.62** | 18.816 | ***188.04*** | 28.290 | 2.7497 | 9.9730 |
| 0.15 | 0.00 | *100.00* | 8.0655 | **264.01** | 19.000 | ***181.33*** | 28.291 | 2.7935 | 9.6382 |
| | 0.01 | *100.00* | 8.0178 | **270.55** | 18.779 | ***187.72*** | 28.217 | 2.7194 | 10.048 |
| | 0.02 | *100.00* | 7.7532 | **263.66** | 18.822 | ***179.38*** | 28.001 | 2.7465 | 9.8450 |
| 0.20 | 0.00 | *100.00* | 7.8426 | **270.20** | 18.723 | ***179.09*** | 27.911 | 2.7358 | 9.8443 |
| | 0.01 | *100.00* | 8.1099 | **266.59** | 18.745 | ***182.57*** | 27.825 | 2.7633 | 9.9745 |
| | 0.02 | *100.00* | 7.7218 | **269.60** | 18.739 | ***179.64*** | 27.819 | 2.7514 | 10.101 |

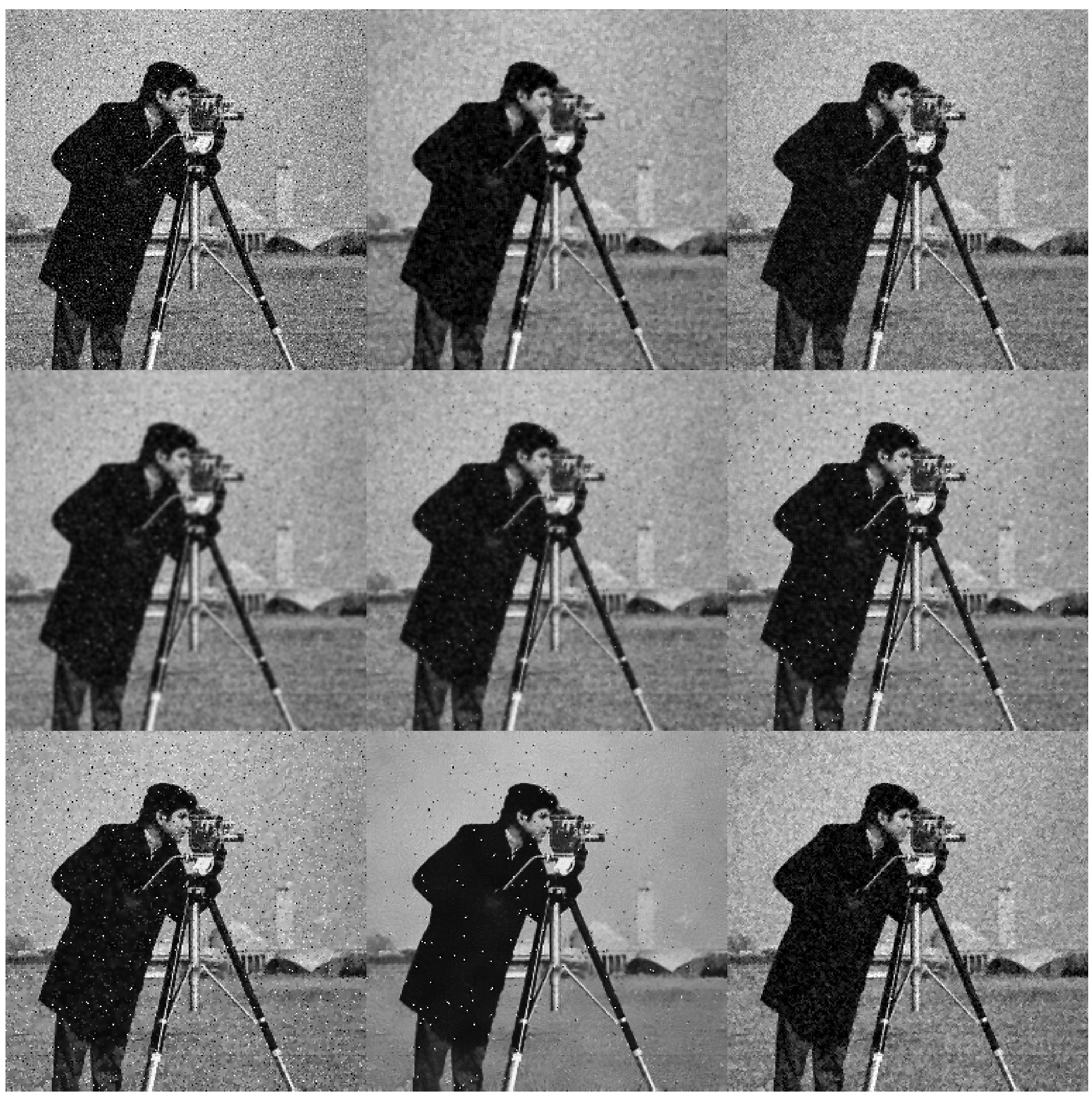

**Fig.6.** Processing results for the “Cameraman” test image corrupted by mixed noise at $\eta = 0.1$ and $\omega = 0.01$: N, M, AM; G, B, W; AD, NLM, and ARC (ordered from left to right, top to bottom)

**Table 2a.** Average $C_{NR}$ (dB) for the noisy "Cameraman" image processed by various filters

| $\eta$ | $\omega$ | M | AM | G | B | W | AD | NLM | ARC |
|---|---|---|---|---|---|---|---|---|---|
| $10^{-3}$ | 0.00 | – 33.941 | **0.0000** | – 34.957 | – 33.413 | – 28.770 | – 29.439 | ***– 17.301*** | *– 27.757* |
| | 0.01 | *2.0488* | **22.366** | 0.6128 | 1.7284 | 0.5176 | – 0.8196 | – 0.1701 | ***6.4298*** |
| | 0.02 | *4.9597* | **22.263** | 3.1558 | 3.9284 | 1.3497 | – 0.4109 | – 0.0908 | ***7.2347*** |
| 0.05 | 0.00 | – 0.0638 | 0.0358 | – 0.4167 | 0.9904 | *3.5975* | ***4.5624*** | **5.7488** | 2.3372 |
| | 0.01 | 3.4582 | *3.6307* | 2.7162 | ***3.7363*** | 2.4292 | 1.4499 | 1.9926 | **5.1625** |
| | 0.02 | *5.3046* | ***5.5592*** | 4.2176 | 4.9208 | 2.4528 | 0.8704 | 1.6816 | **5.9872** |
| 0.10 | 0.00 | 3.7716 | – 0.1060 | 4.4016 | *5.4172* | ***5.7703*** | 2.9553 | **7.6089** | 3.6105 |
| | 0.01 | 4.9278 | 1.1489 | *5.2431* | **5.9716** | 4.5817 | 1.9795 | ***5.8023*** | 4.4849 |
| | 0.02 | *5.7933* | 2.1196 | ***5.8274*** | **6.3106** | 4.1781 | 1.4857 | 5.3437 | 4.9360 |
| 0.15 | 0.00 | 5.1829 | – 0.0507 | *6.4563* | ***6.9227*** | 6.1085 | 1.1121 | **8.3708** | 3.8921 |
| | 0.01 | 5.6878 | 0.5602 | *6.7398* | ***7.0112*** | 5.6017 | 0.9377 | **8.1511** | 4.2413 |
| | 0.02 | 6.1155 | 1.0971 | *6.9612* | ***7.0581*** | 5.3224 | 0.8080 | **8.0401** | 4.4634 |
| 0.20 | 0.00 | 5.7459 | 0.1281 | ***7.4145*** | *7.2625* | 6.1959 | 0.5314 | **8.5953** | 3.9101 |
| | 0.01 | 5.9982 | 0.4887 | ***7.5131*** | *7.2311* | 6.0026 | 0.4817 | **8.6277** | 4.0633 |
| | 0.02 | 6.2238 | 0.8252 | ***7.5940*** | *7.1895* | 5.8716 | 0.4393 | **8.6396** | 4.1669 |

**Table 2b.** Average $C_{VR}$ (dB) for the noisy "Cameraman" image processed by various filters

| $\eta$ | $\omega$ | M | AM | G | B | W | AD | NLM | ARC |
|---|---|---|---|---|---|---|---|---|---|
| $10^{-3}$ | 0.00 | – 38.560 | **0.0000** | – 41.823 | – 40.791 | – 37.320 | – 38.282 | ***– 27.054*** | *– 33.390* |
| | 0.01 | – 10.186 | **22.597** | – 14.419 | – 13.219 | – 11.486 | – 12.031 | *– 7.6852* | ***– 5.4918*** |
| | 0.02 | *– 4.4354* | **24.907** | – 9.4820 | – 8.1741 | – 7.5296 | – 7.9442 | – 5.7511 | ***– 0.4620*** |
| 0.05 | 0.00 | 2.6338 | 0.1438 | 2.4393 | 3.2762 | *4.6364* | ***6.1894*** | **7.1253** | 2.9549 |
| | 0.01 | 3.5182 | 1.1210 | 2.6446 | 3.6032 | *3.9444* | ***5.2385*** | **5.9485** | 3.6871 |
| | 0.02 | *4.2999* | 1.9986 | 2.8291 | 3.8706 | 3.5620 | ***4.5379*** | **5.2961** | 4.2391 |
| 0.10 | 0.00 | 4.8573 | – 0.0581 | 5.8783 | ***6.4135*** | *6.3047* | 4.2399 | **8.4800** | 3.8062 |
| | 0.01 | 5.2471 | 0.4367 | 5.8451 | ***6.4382*** | *5.8545* | 3.8776 | **8.1773** | 4.1065 |
| | 0.02 | 5.6088 | 0.9039 | *5.8257* | ***6.4573*** | 5.5291 | 3.5658 | **8.0254** | 4.3442 |
| 0.15 | 0.00 | 5.6776 | – 0.1064 | *7.1274* | ***7.2587*** | 6.4153 | 1.7684 | **8.8009** | 4.0293 |
| | 0.01 | 5.8935 | 0.2164 | *7.0851* | ***7.2283*** | 6.2459 | 1.6713 | **8.8169** | 4.1792 |
| | 0.02 | 6.0953 | 0.5272 | *7.0425* | ***7.1901*** | 6.0933 | 1.5818 | **8.8225** | 4.3030 |
| 0.20 | 0.00 | 6.0348 | 0.0325 | ***7.6960*** | *7.3335* | 6.3848 | 0.8914 | **8.8121** | 4.0614 |
| | 0.01 | 6.1580 | 0.2648 | ***7.6527*** | *7.2816* | 6.3155 | 0.8537 | **8.8471** | 4.1366 |
| | 0.02 | 6.2780 | 0.4929 | ***7.6130*** | *7.2280* | 6.2495 | 0.8188 | **8.8649** | 4.1997 |

**Table 2c.** Average $C_{SS}$ (SSIM) for the noisy “Cameraman” image processed by various filters

| $\eta$ | $\omega$ | M | AM | G | B | W | AD | NLM | ARC |
|---|---|---|---|---|---|---|---|---|---|
| $10^{-3}$ | 0.00 | 0.8683 | **0.9996** | 0.8603 | 0.8777 | 0.8953 | 0.8729 | ***0.9731*** | *0.9605* |
| | 0.01 | *0.8672* | **0.9985** | 0.7958 | 0.8042 | 0.7363 | 0.6777 | 0.7444 | ***0.9477*** |
| | 0.02 | *0.8659* | **0.9974** | 0.7407 | 0.7407 | 0.6320 | 0.5427 | 0.5892 | ***0.9236*** |
| 0.05 | 0.00 | 0.7198 | 0.5616 | 0.7901 | *0.8074* | 0.7923 | ***0.8631*** | **0.8794** | 0.7002 |
| | 0.01 | 0.7166 | 0.5619 | ***0.7375*** | **0.7450** | 0.6644 | 0.6761 | *0.7180* | 0.6873 |
| | 0.02 | **0.7130** | 0.5623 | ***0.6918*** | *0.6904* | 0.5788 | 0.5453 | 0.6094 | 0.6687 |
| 0.10 | 0.00 | 0.5309 | 0.3380 | *0.6589* | ***0.6677*** | 0.6201 | 0.4783 | **0.7294** | 0.4683 |
| | 0.01 | 0.5267 | 0.3383 | ***0.6245*** | *0.6237* | 0.5440 | 0.4173 | **0.6568** | 0.4590 |
| | 0.02 | 0.5225 | 0.3386 | ***0.5941*** | *0.5847* | 0.4876 | 0.3692 | **0.6040** | 0.4476 |
| 0.15 | 0.00 | 0.4058 | 0.2337 | ***0.5391*** | *0.5278* | 0.4673 | 0.2735 | **0.5935** | 0.3415 |
| | 0.01 | 0.4020 | 0.2338 | ***0.5179*** | *0.4991* | 0.4304 | 0.2553 | **0.5640** | 0.3348 |
| | 0.02 | 0.3981 | 0.2339 | ***0.4983*** | *0.4730* | 0.3996 | 0.2393 | **0.5382** | 0.3274 |
| 0.20 | 0.00 | 0.3244 | 0.1733 | ***0.4481*** | *0.4136* | 0.3667 | 0.1954 | **0.4868** | 0.2652 |
| | 0.01 | 0.3211 | 0.1734 | ***0.4343*** | *0.3949* | 0.3468 | 0.1867 | **0.4712** | 0.2602 |
| | 0.02 | 0.3177 | 0.1734 | ***0.4214*** | *0.3777* | 0.3295 | 0.1786 | **0.4555** | 0.2548 |

**Table 2d.** Average $C_{CE}$ for the noisy “Cameraman” image processed by various filters

| $\eta$ | $\omega$ | M | AM | G | B | W | AD | NLM | ARC |
|---|---|---|---|---|---|---|---|---|---|
| $10^{-3}$ | 0.00 | 100.00 | **2667.9** | ***360.69*** | 27.495 | *178.50* | 26.344 | 2.2111 | 9.1424 |
| | 0.01 | *100.00* | 41.193 | **299.01** | 27.525 | ***178.80*** | 26.318 | 2.4425 | 9.0634 |
| | 0.02 | *100.00* | 38.465 | **283.52** | 28.358 | ***170.25*** | 26.093 | 2.4212 | 9.0593 |
| 0.05 | 0.00 | *100.00* | 24.124 | **311.33** | 32.127 | ***197.20*** | 29.812 | 2.7423 | 10.256 |
| | 0.01 | *100.00* | 23.216 | **282.47** | 32.897 | ***179.61*** | 28.962 | 2.8124 | 10.178 |
| | 0.02 | *100.00* | 22.553 | **291.91** | 32.746 | ***186.99*** | 29.275 | 2.7082 | 10.142 |
| 0.10 | 0.00 | *100.00* | 20.320 | **304.98** | 32.386 | ***191.18*** | 29.135 | 2.6492 | 10.445 |
| | 0.01 | *100.00* | 18.886 | **290.49** | 33.300 | ***187.52*** | 30.028 | 2.7953 | 10.279 |
| | 0.02 | *100.00* | 18.962 | **301.63** | 32.778 | ***190.25*** | 29.893 | 2.7615 | 10.386 |
| 0.15 | 0.00 | *100.00* | 16.672 | **300.79** | 32.367 | ***188.04*** | 29.305 | 2.6480 | 10.460 |
| | 0.01 | *100.00* | 16.002 | **287.07** | 32.796 | ***184.73*** | 29.638 | 2.7144 | 10.424 |
| | 0.02 | *100.00* | 14.553 | **270.57** | 32.603 | ***179.67*** | 29.275 | 2.8525 | 9.7602 |
| 0.20 | 0.00 | *100.00* | 14.631 | **295.57** | 32.290 | ***180.15*** | 29.155 | 2.6849 | 10.337 |
| | 0.01 | *100.00* | 14.349 | **284.23** | 32.557 | ***183.49*** | 29.443 | 2.7284 | 10.263 |
| | 0.02 | *100.00* | 13.527 | **286.73** | 32.659 | ***183.86*** | 29.774 | 2.8678 | 10.171 |

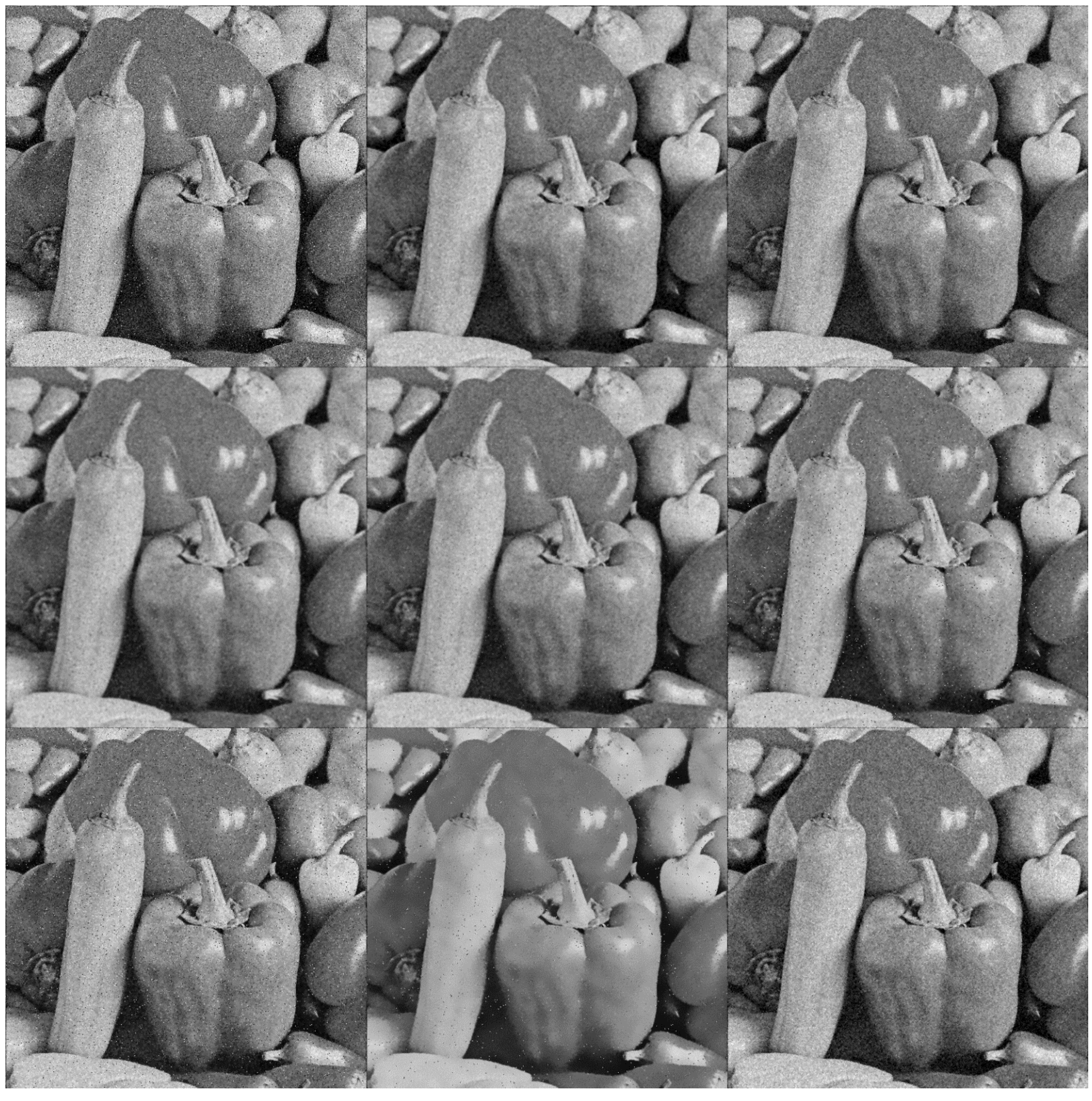

**Fig.7.** Processing results for the "Peppers" test image corrupted by mixed noise at $\eta = 0.1$ and $\omega = 0.01$: N, M, AM; G, B, W; AD, NLM, and ARC (ordered from left to right, top to bottom)

**Table 3a.** Average $C_{NR}$ (dB) for the noisy "Peppers" image processed by various filters

| $\eta$ | $\omega$ | M | AM | G | B | W | AD | NLM | ARC |
|---|---|---|---|---|---|---|---|---|---|
| $10^{-3}$ | 0.00 | – 26.202 | **– 0.7504** | – 28.582 | – 27.827 | – 24.667 | – 27.125 | *– 22.806* | ***– 21.659*** |
| | 0.01 | *9.4135* | **26.518** | 5.5161 | 5.7463 | 0.5937 | – 0.5348 | – 0.2798 | ***10.323*** |
| | 0.02 | ***12.266*** | **26.749** | 7.3088 | 7.2342 | 1.2953 | – 0.2631 | – 0.0226 | *9.5767* |
| 0.05 | 0.00 | 4.8500 | 0.0047 | 5.1271 | 5.6591 | *6.0284* | ***6.6870*** | **7.1244** | 3.8847 |
| | 0.01 | **8.0693** | 3.3460 | *7.2033* | ***7.3317*** | 2.9023 | 1.9668 | 2.5319 | 6.4838 |
| | 0.02 | **9.8145** | 5.2203 | ***8.0724*** | *7.9324* | 2.6866 | 1.1722 | 2.1933 | 7.1117 |
| 0.10 | 0.00 | 6.5852 | 0.0736 | *8.4647* | ***8.5679*** | 6.9977 | 3.3127 | **9.3691** | 4.3652 |
| | 0.01 | 7.5727 | 1.1879 | **8.8288** | ***8.6767*** | 5.4584 | 2.2862 | *7.5926* | 5.1332 |
| | 0.02 | *8.3349* | 2.0823 | **9.0625** | ***8.7047*** | 4.8825 | 1.7402 | 7.2266 | 5.5213 |
| 0.15 | 0.00 | 6.9515 | 0.2273 | ***9.4488*** | *9.0089* | 7.1131 | 1.1662 | **10.202** | 4.3985 |
| | 0.01 | 7.3653 | 0.7679 | ***9.5281*** | *8.9199* | 6.5441 | 1.0009 | **10.079** | 4.6924 |
| | 0.02 | 7.7222 | 1.2538 | ***9.5844*** | *8.8260* | 6.2085 | 0.8733 | **10.003** | 4.8730 |
| 0.20 | 0.00 | 6.9500 | 0.4576 | ***9.7739*** | *8.7028* | 7.1655 | 0.5452 | **10.508** | 4.2998 |
| | 0.01 | 7.1476 | 0.7782 | ***9.7814*** | *8.5972* | 6.9567 | 0.4989 | **10.530** | 4.4200 |
| | 0.02 | 7.3213 | 1.0797 | ***9.7785*** | *8.4888* | 6.8081 | 0.4587 | **10.538** | 4.4965 |

**Table 3b.** Average $C_{VR}$ (dB) for the noisy "Peppers" image processed by various filters

| $\eta$ | $\omega$ | M | AM | G | B | W | AD | NLM | ARC |
|---|---|---|---|---|---|---|---|---|---|
| $10^{-3}$ | 0.00 | – 35.132 | **– 0.1446** | – 37.567 | – 37.259 | – 35.259 | – 37.462 | *– 33.567* | ***– 30.279*** |
| | 0.01 | *– 6.7359* | **22.948** | – 10.735 | – 10.081 | – 9.9496 | – 11.428 | – 9.8973 | ***– 2.4080*** |
| | 0.02 | *– 0.9807* | **25.465** | – 6.2197 | – 5.3585 | – 6.1878 | – 7.4869 | – 6.6488 | ***2.4478*** |
| 0.05 | 0.00 | 5.2276 | 0.0140 | 6.2201 | *6.4430* | 6.3137 | ***7.4135*** | **7.5427** | 4.0520 |
| | 0.01 | 6.0937 | 0.9899 | 6.0390 | **6.4840** | 5.3269 | *6.1895* | ***6.3436*** | 4.7880 |
| | 0.02 | **6.8620** | 1.8691 | *5.9243* | ***6.5141*** | 4.7635 | 5.3082 | 5.7533 | 5.3315 |
| 0.10 | 0.00 | 6.7252 | 0.0914 | ***8.8812*** | *8.8322* | 7.3525 | 4.5254 | **9.8083** | 4.5159 |
| | 0.01 | 7.0838 | 0.5756 | *8.6923* | ***8.7221*** | 7.0031 | 4.1285 | **9.5744** | 4.8021 |
| | 0.02 | 7.4182 | 1.0364 | *8.5369* | ***8.6170*** | 6.6446 | 3.7844 | **9.4899** | 5.0277 |
| 0.15 | 0.00 | 7.0543 | 0.2670 | ***9.6592*** | *9.1529* | 7.5037 | 1.6403 | **10.591** | 4.5574 |
| | 0.01 | 7.2420 | 0.5806 | ***9.5475*** | *9.0454* | 7.3484 | 1.5484 | **10.635** | 4.6945 |
| | 0.02 | 7.4195 | 0.8843 | ***9.4485*** | *8.9419* | 7.1916 | 1.4631 | **10.666** | 4.8056 |
| 0.20 | 0.00 | 7.0739 | 0.5362 | ***9.9451*** | *8.8611* | 7.5740 | 0.7449 | **10.865** | 4.4794 |
| | 0.01 | 7.1780 | 0.7619 | ***9.8736*** | *8.7664* | 7.4937 | 0.7127 | **10.899** | 4.5458 |
| | 0.02 | 7.2733 | 0.9829 | ***9.8040*** | *8.6701* | 7.4164 | 0.6819 | **10.919** | 4.5984 |

**Table 3c.** Average $C_{SS}$ (SSIM) for the noisy "Peppers" image processed by various filters

| $\eta$ | $\omega$ | M | AM | G | B | W | AD | NLM | ARC |
|---|---|---|---|---|---|---|---|---|---|
| $10^{-3}$ | 0.00 | 0.8853 | **0.9996** | 0.8819 | 0.8852 | 0.8949 | 0.8590 | *0.9201* | ***0.9612*** |
| | 0.01 | *0.8844* | **0.9978** | 0.8218 | 0.8188 | 0.7050 | 0.6494 | 0.6796 | ***0.9500*** |
| | 0.02 | *0.8834* | **0.9959** | 0.7696 | 0.7595 | 0.5886 | 0.5045 | 0.5371 | ***0.9259*** |
| 0.05 | 0.00 | 0.7474 | 0.5151 | 0.8155 | *0.8188* | 0.7947 | ***0.8380*** | **0.8435** | 0.6961 |
| | 0.01 | *0.7444* | 0.5157 | **0.7656** | ***0.7608*** | 0.6426 | 0.6382 | 0.6783 | 0.6837 |
| | 0.02 | **0.7411** | 0.5163 | ***0.7216*** | *0.7087* | 0.5450 | 0.4991 | 0.5739 | 0.6642 |
| 0.10 | 0.00 | 0.5470 | 0.2616 | ***0.6812*** | *0.6735* | 0.5824 | 0.4051 | **0.7035** | 0.4324 |
| | 0.01 | 0.5424 | 0.2623 | **0.6484** | *0.6313* | 0.5144 | 0.3452 | ***0.6417*** | 0.4229 |
| | 0.02 | 0.5377 | 0.2629 | **0.6188** | *0.5929* | 0.4575 | 0.2980 | ***0.5970*** | 0.4111 |
| 0.15 | 0.00 | 0.4016 | 0.1625 | ***0.5546*** | *0.5190* | 0.4245 | 0.1925 | **0.5712** | 0.2897 |
| | 0.01 | 0.3974 | 0.1629 | ***0.5343*** | *0.4910* | 0.3934 | 0.1774 | **0.5495** | 0.2831 |
| | 0.02 | 0.3930 | 0.1633 | ***0.5154*** | *0.4650* | 0.3657 | 0.1641 | **0.5291** | 0.2758 |
| 0.20 | 0.00 | 0.3058 | 0.1127 | ***0.4566*** | *0.3891* | 0.3273 | 0.1224 | **0.4662** | 0.2090 |
| | 0.01 | 0.3021 | 0.1130 | ***0.4436*** | *0.3711* | 0.3106 | 0.1160 | **0.4540** | 0.2043 |
| | 0.02 | 0.2983 | 0.1133 | ***0.4312*** | *0.3541* | 0.2956 | 0.1101 | **0.4416** | 0.1992 |

**Table 3d.** Average $C_{CE}$ for the noisy "Peppers" image processed by various filters

| $\eta$ | $\omega$ | M | AM | G | B | W | AD | NLM | ARC |
|---|---|---|---|---|---|---|---|---|---|
| $10^{-3}$ | 0.00 | *100.00* | 38.871 | **375.29** | 34.986 | ***137.16*** | 34.372 | 3.2966 | 7.7216 |
| | 0.01 | *100.00* | 37.478 | **376.90** | 33.607 | ***136.63*** | 33.184 | 3.1859 | 7.6921 |
| | 0.02 | *100.00* | 37.444 | **392.33** | 34.411 | ***139.43*** | 33.317 | 3.2355 | 7.8278 |
| 0.05 | 0.00 | *100.00* | 23.521 | **412.83** | 38.036 | ***143.85*** | 35.843 | 3.4149 | 8.5202 |
| | 0.01 | *100.00* | 21.407 | **390.34** | 36.133 | ***140.04*** | 35.564 | 3.3687 | 8.0039 |
| | 0.02 | *100.00* | 20.771 | **402.13** | 37.339 | ***143.78*** | 34.214 | 3.2645 | 8.1024 |
| 0.10 | 0.00 | *100.00* | 19.121 | **381.89** | 38.127 | ***134.43*** | 35.846 | 3.4614 | 8.0709 |
| | 0.01 | *100.00* | 19.223 | **404.07** | 37.263 | ***143.02*** | 36.091 | 3.3995 | 8.2725 |
| | 0.02 | *100.00* | 17.986 | **394.94** | 38.968 | ***137.20*** | 36.127 | 3.4735 | 7.9967 |
| 0.15 | 0.00 | *100.00* | 17.015 | **403.05** | 40.036 | ***139.74*** | 36.814 | 3.4582 | 8.2940 |
| | 0.01 | *100.00* | 16.130 | **404.76** | 37.014 | ***141.96*** | 35.398 | 3.3612 | 8.2411 |
| | 0.02 | *100.00* | 15.910 | **404.82** | 38.143 | ***142.51*** | 36.001 | 3.4365 | 8.3782 |
| 0.20 | 0.00 | *100.00* | 14.540 | **405.19** | 37.966 | ***136.58*** | 35.520 | 3.4204 | 8.3223 |
| | 0.01 | *100.00* | 13.067 | **382.26** | 37.244 | ***130.09*** | 35.828 | 3.4442 | 7.8008 |
| | 0.02 | *100.00* | 13.116 | **371.33** | 38.423 | ***128.84*** | 35.285 | 3.4728 | 7.8356 |

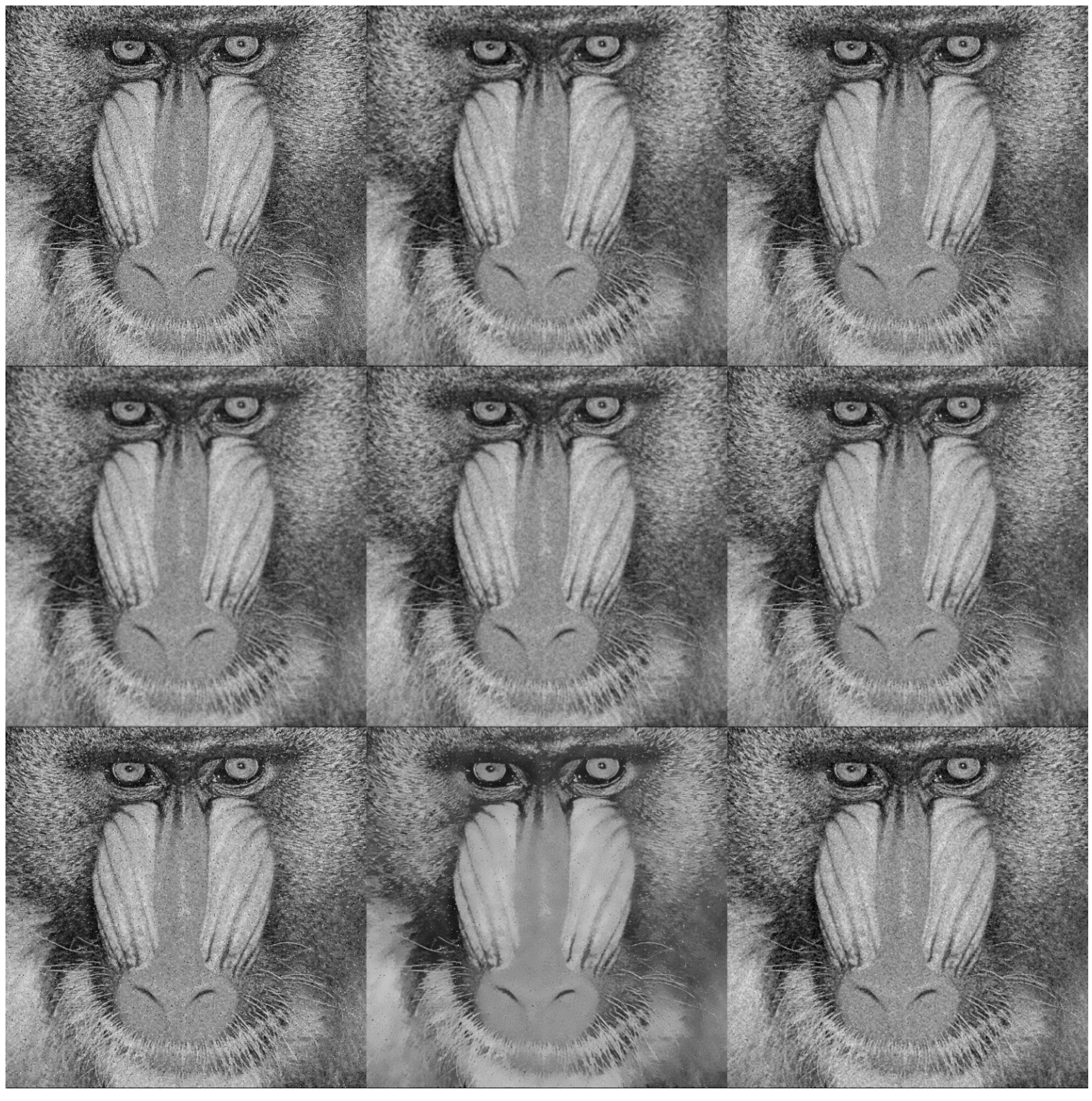

**Fig.8.** Processing results for the "Baboon" test image corrupted by mixed noise at $\eta = 0.1$ and $\omega = 0.01$: N, M, AM; G, B, W; AD, NLM, and ARC (ordered from left to right, top to bottom)

**Table 4a.** Average $C_{NR}$ (dB) for the noisy "Baboon" image processed by various filters

| $\eta$ | $\omega$ | M | AM | G | B | W | AD | NLM | ARC |
|---|---|---|---|---|---|---|---|---|---|
| $10^{-3}$ | 0.00 | – 37.661 | **– 22.485** | – 37.549 | – 36.930 | – 33.777 | – 33.125 | ***– 25.649*** | *– 32.183* |
| | 0.01 | – 2.1073 | **11.887** | – 2.2045 | – 1.6290 | – 1.4526 | – 1.9039 | *– 0.6508* | ***2.6600*** |
| | 0.02 | 0.8553 | **13.911** | 0.5619 | *1.0857* | 0.0223 | – 1.0294 | – 0.2768 | ***4.5869*** |
| 0.05 | 0.00 | – 3.0516 | – 0.1075 | – 2.6320 | – 1.9968 | – 0.0084 | ***1.1173*** | **2.4634** | *0.3204* |
| | 0.01 | 0.1416 | ***3.0839*** | 0.3648 | 0.9445 | 0.8145 | 0.5070 | *1.3459* | **3.0859** |
| | 0.02 | 1.9211 | **4.8626** | 1.9611 | *2.4782* | 1.3504 | 0.3337 | 1.4749 | ***4.2296*** |
| 0.10 | 0.00 | 1.8276 | 0.0135 | 2.7876 | *3.3963* | ***3.7472*** | 1.8544 | **4.7341** | 2.8605 |
| | 0.01 | 2.8171 | 1.0659 | *3.6207* | ***4.1466*** | 3.5071 | 1.3652 | **4.6326** | 3.6174 |
| | 0.02 | 3.6001 | 1.9149 | *4.2576* | ***4.6980*** | 3.4864 | 1.0764 | **4.7684** | 4.0815 |
| 0.15 | 0.00 | 3.9341 | 0.1910 | *5.3998* | ***5.7750*** | 5.1894 | 0.8460 | **6.1371** | 3.5887 |
| | 0.01 | 4.3561 | 0.6966 | *5.7142* | ***5.9887*** | 5.0424 | 0.7372 | **6.2766** | 3.8654 |
| | 0.02 | 4.7300 | 1.1536 | *5.9849* | ***6.1589*** | 4.9783 | 0.6508 | **6.4133** | 4.0576 |
| 0.20 | 0.00 | 4.9069 | 0.4515 | ***6.8060*** | *6.6956* | 5.8954 | 0.4414 | **7.1076** | 3.7915 |
| | 0.01 | 5.1134 | 0.7508 | ***6.9452*** | *6.7352* | 5.8439 | 0.4072 | **7.2111** | 3.9004 |
| | 0.02 | 5.3012 | 1.0316 | ***7.0682*** | *6.7600* | 5.8161 | 0.3771 | **7.3056** | 3.9773 |

**Table 4b.** Average $C_{VR}$ (dB) for the noisy "Baboon" image processed by various filters

| $\eta$ | $\omega$ | M | AM | G | B | W | AD | NLM | ARC |
|---|---|---|---|---|---|---|---|---|---|
| $10^{-3}$ | 0.00 | – 46.234 | **– 7.1109** | – 47.337 | – 46.852 | – 44.232 | – 43.870 | ***– 35.435*** | *– 40.535* |
| | 0.01 | – 17.830 | **15.277** | – 19.242 | – 18.678 | – 17.297 | – 16.602 | *– 12.468* | ***– 12.403*** |
| | 0.02 | – 12.022 | **17.565** | – 13.718 | – 13.093 | – 12.503 | – 11.787 | *– 9.6270* | ***– 6.9821*** |
| 0.05 | 0.00 | – 1.7295 | – 0.0291 | – 1.5977 | – 1.0943 | 0.5039 | ***1.6154*** | **2.8295** | *0.8171* |
| | 0.01 | – 0.8285 | 0.8922 | – 0.9683 | – 0.4066 | 0.2862 | *1.4374* | **2.0717** | ***1.5656*** |
| | 0.02 | – 0.0246 | ***1.7159*** | – 0.4145 | 0.1897 | 0.2819 | 1.2958 | *1.6745* | **2.1593** |
| 0.10 | 0.00 | 2.5127 | 0.0320 | 3.4894 | *3.9652* | ***4.1638*** | 2.4187 | **5.1990** | 3.0388 |
| | 0.01 | 2.9033 | 0.4916 | 3.6880 | ***4.1765*** | *3.9348* | 2.2307 | **5.0649** | 3.3228 |
| | 0.02 | 3.2711 | 0.9287 | *3.8748* | ***4.3670*** | 3.7949 | 2.0658 | **5.0184** | 3.5586 |
| 0.15 | 0.00 | 4.3381 | 0.2127 | *5.8779* | ***6.1235*** | 5.5369 | 1.1155 | **6.5986** | 3.7605 |
| | 0.01 | 4.5491 | 0.5109 | *5.9635* | ***6.1833*** | 5.4328 | 1.0557 | **6.6362** | 3.8931 |
| | 0.02 | 4.7529 | 0.8006 | *6.0477* | ***6.2374*** | 5.3568 | 0.9996 | **6.6807** | 4.0050 |
| 0.20 | 0.00 | 5.2190 | 0.5136 | ***7.1821*** | *6.9684* | 6.2410 | 0.5658 | **7.5606** | 4.0040 |
| | 0.01 | 5.3398 | 0.7294 | ***7.2277*** | *6.9674* | 6.2017 | 0.5417 | **7.6132** | 4.0667 |
| | 0.02 | 5.4540 | 0.9394 | ***7.2683*** | *6.9595* | 6.1693 | 0.5184 | **7.6623** | 4.1176 |

**Table 4c.** Average $C_{SS}$ (SSIM) for the noisy "Baboon" image processed by various filters

| $\eta$ | $\omega$ | M | AM | G | B | W | AD | NLM | ARC |
|---|---|---|---|---|---|---|---|---|---|
| $10^{-3}$ | 0.00 | 0.7212 | **0.9981** | 0.7135 | 0.7399 | 0.8231 | 0.8085 | ***0.9290*** | *0.9237* |
| | 0.01 | 0.7194 | **0.9960** | 0.6855 | 0.7112 | 0.7158 | 0.6891 | *0.7742* | ***0.9130*** |
| | 0.02 | *0.7175* | **0.9938** | 0.6600 | 0.6844 | 0.6414 | 0.5986 | 0.6657 | ***0.8946*** |
| 0.05 | 0.00 | 0.6518 | 0.7562 | 0.6856 | 0.7154 | 0.7713 | ***0.8178*** | **0.8500** | *0.7919* |
| | 0.01 | 0.6493 | ***0.7553*** | 0.6609 | 0.6890 | 0.6802 | 0.7001 | *0.7334* | **0.7805** |
| | 0.02 | 0.6466 | ***0.7542*** | 0.6385 | *0.6643* | 0.6163 | 0.6111 | 0.6537 | **0.7642** |
| 0.10 | 0.00 | 0.5428 | 0.5259 | 0.6230 | *0.6560* | ***0.6643*** | 0.5990 | **0.7113** | 0.6163 |
| | 0.01 | 0.5394 | 0.5255 | 0.6047 | ***0.6342*** | 0.6055 | 0.5448 | **0.6619** | *0.6059* |
| | 0.02 | 0.5361 | 0.5252 | 0.5877 | ***0.6135*** | 0.5604 | 0.4991 | **0.6233** | *0.5933* |
| 0.15 | 0.00 | 0.4480 | 0.3775 | *0.5533* | ***0.5789*** | 0.5508 | 0.4056 | **0.5837** | 0.4829 |
| | 0.01 | 0.4445 | 0.3775 | *0.5403* | ***0.5617*** | 0.5192 | 0.3830 | **0.5620** | 0.4742 |
| | 0.02 | 0.4410 | 0.3773 | *0.5280* | **0.5451** | 0.4923 | 0.3624 | ***0.5418*** | 0.4645 |
| 0.20 | 0.00 | 0.3730 | 0.2777 | ***0.4892*** | **0.4966** | 0.4589 | 0.2965 | *0.4837* | 0.3851 |
| | 0.01 | 0.3694 | 0.2777 | ***0.4796*** | **0.4831** | 0.4412 | 0.2844 | *0.4707* | 0.3779 |
| | 0.02 | 0.3660 | 0.2776 | **0.4703** | ***0.4700*** | 0.4253 | 0.2730 | *0.4582* | 0.3702 |

**Table 4d.** Average $C_{CE}$ for the noisy "Baboon" image processed by various filters

| $\eta$ | $\omega$ | M | AM | G | B | W | AD | NLM | ARC |
|---|---|---|---|---|---|---|---|---|---|
| $10^{-3}$ | 0.00 | *100.00* | 36.330 | **370.07** | 35.312 | ***133.10*** | 33.539 | 3.2530 | 7.5272 |
| | 0.01 | *100.00* | 21.927 | **373.54** | 33.626 | ***129.15*** | 31.159 | 2.9526 | 7.2855 |
| | 0.02 | *100.00* | 20.193 | **364.26** | 29.059 | ***137.46*** | 34.067 | 3.0960 | 7.4996 |
| 0.05 | 0.00 | *100.00* | 40.395 | **424.33** | 38.432 | ***145.24*** | 34.951 | 3.1345 | 8.2789 |
| | 0.01 | *100.00* | 27.599 | **378.20** | 35.627 | ***132.31*** | 32.288 | 3.0435 | 7.5938 |
| | 0.02 | *100.00* | 26.392 | **378.75** | 39.122 | ***139.19*** | 36.501 | 3.2727 | 8.1571 |
| 0.10 | 0.00 | *100.00* | 26.333 | **401.66** | 37.344 | ***135.20*** | 33.837 | 3.1891 | 8.0282 |
| | 0.01 | *100.00* | 23.276 | **390.41** | 37.000 | ***134.71*** | 34.064 | 3.1844 | 7.9118 |
| | 0.02 | *100.00* | 22.537 | **400.91** | 40.119 | ***141.12*** | 37.474 | 3.4518 | 8.3231 |
| 0.15 | 0.00 | *100.00* | 19.700 | **404.43** | 37.483 | ***133.39*** | 33.824 | 3.2154 | 7.8454 |
| | 0.01 | *100.00* | 19.546 | **392.08** | 38.233 | ***132.98*** | 34.451 | 3.2551 | 7.9395 |
| | 0.02 | *100.00* | 20.196 | **402.44** | 40.048 | ***137.16*** | 37.534 | 3.4832 | 8.2622 |
| 0.20 | 0.00 | *100.00* | 18.245 | **387.96** | 36.787 | ***135.03*** | 33.629 | 3.2055 | 8.0355 |
| | 0.01 | *100.00* | 18.511 | **399.37** | 36.599 | ***137.59*** | 35.127 | 3.2081 | 8.0623 |
| | 0.02 | *100.00* | 18.620 | **406.66** | 39.351 | ***139.50*** | 37.644 | 3.4029 | 8.4604 |

In the above tables, the three best results for each metric are highlighted in **bold**, ***bold-italic***, and *italic* types, in descending order of performance. According to the obtained results, the most computationally efficient local filters under the considered parameter settings are the Gaussian, Wiener, and median filters. The least efficient among the evaluated methods, the NLM-filter, is approximately thirty times slower than the median filter, despite the relatively small sizes of both the search and similarity windows. The ARC-filter is more efficient than the NLM-filter but is inferior in this parameter to the median filter, being slower than the latter by more than an order of magnitude.

Regarding the image restoration quality under the considered noise parameters, the following observations can be made. As expected, the NLM-filter globally provides the highest processing quality and the minimum number of artifacts. Nevertheless, starting from a certain level of additive Gaussian noise, the image processed by this filter begins to look overly “synthetic” (“plastic”) compared to the original. Other filters exhibit their best results specifically under the most favorable conditions characteristic of their design. In particular, if the image lacks high-contrast fine details (at the Nyquist frequency level)—meaning the image features a sufficiently smooth spatial intensity transition from pixel to pixel—the AM-filter and the median filter perform best when impulse noise dominates. When the two types of noise compete, the highest processing quality is achieved using the Wiener filter and the anisotropic diffusion algorithm. Conversely, when additive Gaussian noise dominates, the bilateral and Gaussian filters expectedly yield the best results. The ARC-filter demonstrates a highly stable, above-average performance regardless of the noise levels within the specified noise ranges. However, if the image contains numerous fine, high-contrast details (as shown in Fig.1), the ARC-filter globally achieves the highest metrics among all local methods, excluding the NLM-filter. Depending on the noise levels and their balance, the AD-filter, the bilateral filter, and the Wiener filter follow in terms of processing quality. Regarding the generation of processing artifacts, the ARC-filter exhibits a similarly low error rate as the NLM-filter, significantly outperforming all other filters in this aspect. Furthermore, the ARC-filter is completely free from the “plastic” artifact characteristic of the NLM-filter; the resulting image maintains a natural appearance, albeit visually slightly noisier.

Based on visual assessment, the closest analog to image processing with the ARC-filter is adaptive median filtration. This is to be expected, as both methods truncate the tails of the pixel intensity distribution within the sliding window relative to a certain baseline value characteristic of the statistical majority. The higher selectivity and superior image restoration metrics achieved by the ARC-filter compared to the AM-filter at the considered low impulse noise levels are explained by the fact that the ARC-filter algorithm accounts for the possibility of both unimodal and bimodal pixel intensity distributions within the sliding window.

## Conclusion

An evaluation of the ARC-filter performance across the considered ranges of additive Gaussian and impulse noise levels allows for the following conclusions:

● Metric-based analysis demonstrates a fundamental advantage of the developed filter when processing images with gradient backgrounds and regular structures at the Nyquist frequency level. Under pronounced mixed noise conditions, where traditional methods and algorithms irreversibly blur or eliminate fine lines, the ARC-filter precisely preserves object topology, silhouette sharpness, and the contrast of extended boundaries due to its adaptation mechanism to local bimodal intensity distributions.

● It has been experimentally verified that under combined noise conditions, the ARC-filter generates virtually no internal artifacts (such as false contouring or blocking) and is completely free from the visual “plastic” effect characteristic of the NLM-filter. Regardless of the noise level, the ARC-filter demonstrates high robustness, maintaining a stable performance plateau and preventing the edge degradation typical of other filters. Furthermore, in the region of vanishingly small noise, the NLM-filter and ARC-filter algorithms exhibit very close metric values.

● The capacity of the developed algorithm to efficiently suppress spot impulse outliers and Gaussian noise tails, while strictly preserving the natural background micro-texture and fine details, defines its primary area of application. This lies in the field of digital medical imaging (such as radiography, CT, MRI, ultrasound, etc.). On digital diagnostic images, the ARC-filter could efficiently remove hardware-induced noise while eliminating the risk of blurring or losing small pathological lesions, fibrous structures, or anomaly contours, which is critically important for preventing false-negative diagnoses.

### Declaration of AI and AI-assisted technologies

The "Gemini 3.6 Flash" multimodal AI model by Google was utilized during the development phase of this work. Specifically, AI technologies were applied to solve the following tasks:

- Developing a software MATLAB script to generate a synthetic test image.
- Developing executable software scripts implementing the ARC-filter algorithm.
- Performing deep optimization of the ARC-filter code to reduce image processing time.
- Drafting the initial translation of the manuscript text from Russian into English.

Algorithm testing and interpretation of the results were performed entirely by the author.

The MATLAB and Python scripts of the proposed ARC-filter are publicly available in the GitHub repository under the free MIT license, and are also archived in the Zenodo repository with an assigned DOI.